\documentclass[symmetric,justified]{tufte-book}
\hypersetup{colorlinks}

\title{A Framework \LARGE for Searching for\Huge\break General Artificial Intelligence}
\pagetitle{A Framework for Searching for General AI}
\author[Marek Rosa, Jan Feyereisl]{M. Rosa, J. Feyereisl \& The GoodAI Team}
\publisher{Version 1}

\usepackage{enumitem}

\usepackage{lipsum}

\usepackage{booktabs}

\usepackage{graphicx}
\setkeys{Gin}{width=\linewidth,totalheight=\textheight,keepaspectratio}
\graphicspath{{graphics/}}

\usepackage{fancyvrb}
\fvset{fontsize=\normalsize}

\usepackage{xspace}

\newcommand{\monthyear}{%
  \ifcase\month\or January\or February\or March\or April\or May\or June\or
  July\or August\or September\or October\or November\or
  December\fi\space\number\year
}

\usepackage{units}

\newcommand{\hlred}[1]{\textcolor{Maroon}{#1}}
\newcommand{\hangleft}[1]{\makebox[0pt][r]{#1}}

\providecommand{\XeLaTeX}{X\lower.5ex\hbox{\kern-0.15em\reflectbox{E}}\kern-0.1em\LaTeX}

\newcommand{\tuftebs}{\symbol{'134}}
\newcommand{\doccmddef}[2][]{%
  \hlred{\texttt{\tuftebs#2}}\label{cmd:#2}%
  \ifthenelse{\isempty{#1}}%
    {
      \index{#2 command@\protect\hangleft{\texttt{\tuftebs}}\texttt{#2}}
    }%
    {
      \index{#2 command@\protect\hangleft{\texttt{\tuftebs}}\texttt{#2} (\texttt{#1} package)}
      \index{#1 package@\texttt{#1} package}\index{packages!#1@\texttt{#1}}
    }%
}
\newcommand{\doccmd}[2][]{%
  \texttt{\tuftebs#2}%
  \ifthenelse{\isempty{#1}}%
    {
      \index{#2 command@\protect\hangleft{\texttt{\tuftebs}}\texttt{#2}}
    }%
    {
      \index{#2 command@\protect\hangleft{\texttt{\tuftebs}}\texttt{#2} (\texttt{#1} package)}
      \index{#1 package@\texttt{#1} package}\index{packages!#1@\texttt{#1}}
    }%
}

\usepackage{makeidx}
\makeindex

\begin{document}

\frontmatter



\maketitle

\newpage
\begin{fullwidth}

\begin{table}[h]
\small
  \begin{center}
    \begin{tabular}{r p{13cm}}
  	Versions	& \\ \midrule
    \textbf{1.0}& \textbf{current version}: initial release for the general public
    \newline \\ 
    2.0			& \textit{planned future version}: additional illustrations demonstrating framework principles
    \newline \\ 
    3.0			& \textit{planned future version}: additional mathematical formalizations and proofs
    \newline \\ 
	\end{tabular}
\end{center}
\end{table}

~\vfill

\thispagestyle{empty}
\setlength{\parindent}{0pt}
\setlength{\parskip}{\baselineskip}
\large 
\textsc{Correspondence}: \texttt{\{marek.rosa,jan.feyereisl\}@goodai.com} \newline
\normalsize

Copyright \copyright\ \the\year\ GoodAI


\par\smallcaps{www.GoodAI.com}

\par The \LaTeX~template used in this document is licensed under the Apache License, Version 2.0 (the ``License''); you may not use this file except in compliance with the License. You may obtain a copy
of the License at \url{http://www.apache.org/licenses/LICENSE-2.0}. Unless
required by applicable law or agreed to in writing, software distributed
under the License is distributed on an \smallcaps{``AS IS'' BASIS, WITHOUT
WARRANTIES OR CONDITIONS OF ANY KIND}, either express or implied. See the
License for the specific language governing permissions and limitations
under the License.\index{license}

\par\textit{First printing, \monthyear}
\end{fullwidth}





\mainmatter

\cleardoublepage

\chapter*{Abstract}
\label{ch:abstract}

There is a significant lack of unified approaches to building generally intelligent machines. The majority of current artificial intelligence research operates within a very narrow field of focus, frequently without considering the importance of the 'big picture'. In this document, we seek to describe and unify principles that guide the basis of our development of general artificial intelligence. These principles revolve around the idea that intelligence is a tool for searching for general solutions to problems. We define intelligence as the ability to acquire skills that narrow this search, diversify it and help steer it to more promising areas. We also provide suggestions for studying, measuring, and testing the various skills and abilities that a human-level\footnote{Despite its vagueness, and the lack of a universal definition, the term provides a notion of an ability of a machine to solve as many tasks to at least the same level of accuracy as most humans.} intelligent machine needs to acquire. The document aims to be both implementation agnostic, and to provide an analytic, systematic, and scalable way to generate hypotheses that we believe are needed to meet the necessary conditions in the search for general artificial intelligence. We believe that such a framework is an important stepping stone for bringing together definitions, highlighting open problems, connecting researchers willing to collaborate, and for unifying the arguably most significant search of this century.

\chapter{Introduction}
\label{ch:introduction}

The search for general artificial intelligence (AI) is one of the biggest challenges of this century\cite{Lake2016,Mikolov2015,Hutter2005,Legg2007-md}. In this document we seek to describe and unify the main principles behind our approach to solving this enormous challenge. 

\newthought{We believe} that to tackle such a challenge, one can first begin with a certain theory or a collection of ideas and beliefs that are more or less correct, yet can be somewhat vague, not well defined and possibly puzzling in some ways. Then, one can work backwards from those concepts, refine those ideas, progressively eliminating vagueness and eventually providing robust and clear definitions, ideally crystallizing in the main minimal tenets of the true underlying theory. These are in fact the \textit{analytic} and \textit{constructive} stages of Bertrand Russel's philosophical method\cite{Klement2013}, respectively. This document should currently be viewed as being in the early \textit{analytic} stage and a work in progress. With time however, it will be continually refined, and eventually, we believe, developed into a robust framework for building intelligent machines, especially with the help of the community.

\newthought{We start} with the reasoning behind the creation of this framework, followed by as of yet informal definitions (cf. tables of definitions in the Appendix) of our understanding of intelligence and related concepts. We then propose a way to build and educate general AI systems quickly and effectively through gradual and guided learning. We conclude with a critical look at our proposal and identification of important next steps. We start with the very basics, the principles and sometimes even describe the obvious. This is however necessary, in order to unify the definitions and approaches and subsequently the thinking of the community interested in collaborating and allow for more rapid and better-defined progress, enhanced collaboration, and reduction of the search space for general artificial intelligence.

\section{What is the goal of this framework?}
\label{sec:goal}

This framework will eventually provide a unified collection of principles, ideas, definitions and formalizations of our thoughts on the process of developing general artificial intelligence. It allows us to bring together all that we believe is important for defining a basis on which we and others can build. It will act as a common language that everyone can understand, and provide a starting point for a platform for further discussions and evolution of our ideas.

\newthought{Within this document}, we also provide a list of ``next steps'': important research topics that we believe the community as a whole should focus on next in order to allow for significant progress in this field. With a separable definition of the problems we are tackling, the work can be split among various research groups (both internally as well as among external collaborators, academia, students, other research centers and hobbyists). This framework does not focus on narrow artificial intelligence (which solves very specific problems well), or on short-term commercialization. The aim is long-term, with possible exploitation of useful applications along the way.

\section{Who is this framework for?}
\label{sec:forwho}

This framework is for both a general and a technical audience. The aim is to make it accessible to everyone first, yet with enough detail eventually, that advanced readers should also benefit from it. No previous experience with AI or robotics is necessary at this stage. If something is not clear in this document, it means that we have failed finding an appropriate exposition for presenting our ideas. Please let us know so we can deliver a better explanation in the next edition.

\section{What is general AI and how can it be useful?}
\label{sec:generalai}

We view artificial intelligence (AI) systems as programs that are able to learn, adapt, be creative and solve problems. Some divide the field into narrow (weak) and general (strong) AI\cite[-4cm]{Adams2012-qg,McCarthy1987-cl,Goertzel2007-ih,McCarthy1969-xg}. This somewhat contrived division primarily highlights the difference in universality of the underlying technology. While narrow AI is usually able to solve only one specific problem and is unable to transfer skills from domain to domain, general AI aims for a human-level skill set\cite{Rajani2011}.

\newthought{No one has developed} a practical general AI yet. With general AI, humankind will be able do many things we simply cannot do with our current level of technology. We will automate science, engineering, production, manufacturing, robots, entertainment, anything you can think of, and more. We believe that general AI will help us become better people, augment our own intelligence, and recursively self-improve.

\chapter{Foundations of general AI}
\label{ch:foundations}

Building general artificial intelligence is a highly complex process, comprising of a large variety of heterogeneous challenges. Over the years, a variety of disparate ideas and definitions on how to tackle this process emerged\cite[-2cm]{Hutter2005,Gershman2015,Legg2007-vl,Legg2007-md,Lake2016}. To clarify and unify our thoughts and ideas that govern the foundations and basis upon which we build, we first provide descriptions of the two core principles underlying our general AI development.

\section{What is intelligence?}
\label{sec:intelligence}

We view intelligence as a problem solving tool that searches for solutions to problems in dynamic, complex and uncertain environments. This view, consistent with numerous existing perspectives 
\citep{Hutter2005,Lake2016,Legg2007-vl,Marblestone2016-zk}
, can be simplified even further: from a computational point of view, most solvable problems can be viewed as search and optimization problems\cite{Polya2014-vb,Minsky1961-lv}, and the goal of intelligence (or an intelligent agent) is to always find the best available solutions with as few resources and as quickly as possible \citep{Gershman2015,Marblestone2016-zk}.

\newthought{We believe} that intelligence can achieve this by discovering skills (abilities, heuristics, shortcuts, tricks) that narrow the search space\cite{Lenat1991-rx}, diversify it, and efficiently help steer it towards areas that are potentially more promising\cite{Bengio2014-tz}.
\newline

We argue that some of the most useful skills are the \textbf{capacity to gradually acquire new skills} - which helps in \textbf{exploiting accumulated knowledge} in order to speed up the acquisition of additional skills, the reuse of existing skills, and recursive self-improvement\cite{Steunebrink2016-dd,Nivel2013}. This way, the intelligent agent continually creates a repertoire of skills that are, essentially, building blocks for new, more complex skills.
\newline

The possibly limitless accumulation of skills in an intelligent machine is naturally bounded by limited resources (time, memory, atoms, energy, etc.). This additional constraint on intelligence results in favoring ``efficient'' skills and operation under bounded rationality \citep{Gershman2015}.

\newthought{In addition} to gradual learning, guided learning helps narrow the search, because at each step, an intelligent teacher guides the agent and hence the agent has to search for a new solution only within a small and useful area\cite{Schmidhuber2013-hr,Bengio2009-fd,Gulcehre2016-ht,Krueger2009-vv,Vapnik2015a}, decreasing the number of candidate solutions, and thereby reducing the complexity of the search space \citep{Alexander2015-mx,Amodei2016-qf,Zhang2016-ug,Zaremba2015-ia,Zaremba2014-co,Gulcehre2016-mb}. On the other hand, without gradual or guided learning, if the agent is expected to find a solution to a complex problem too far from its current capabilities, it might never find a solution within any reasonable amount of time.

\section{What is a skill?}
\label{sec:skill}

A skill is any assumption about a problem that narrows and diversifies the search for a solution and points the search towards more promising areas. It is not guaranteed to be perfect, but sufficient to meet immediate goals, i.e. progress towards ideally a general solution. It can be thought of as akin to a continuum between Minsky's notion of mental skills in his Society of Mind\cite{Minsky1986-as} and Orallo's cognitive abilities\cite{Hernandez-Orallo2016}.

\newthought{For the purpose} of this framework, we see the following words as synonyms for a ``skill'': \textit{ability, heuristic, behavior, strategy, solution, algorithm, shortcut, trick, approximation, exploiting structure in data, and more}. This covers a wide variety of methods for helping to solve problems and might seem counter-intuitive at first. Nonetheless, our aim is to give a clear exposition of the generality of the notion of a ``skill'' as a problem-solving mechanism and a device for narrowing and diversifying the search for solutions from a variety of viewpoints.
\newline

A skill can also be considered a bias. It constrains the search space or restricts behavior. 
\newline 

Some skills can be general, yet simple (e.g. the ability to detect a simple pattern such as a line or edge) while others can be general and complex (e.g. the ability to navigate through an environment, the ability to understand and use language, etc.). 

\newthought{One possible way} to compare the intelligence of various agents is to measure and compare the complexity and generality of problems they can solve\cite[-2cm]{Hernandez-Orallo1998}. Another way can be in terms of measuring the effectiveness as well as efficiency of their skills \citep{Gershman2015} in terms of their sample complexity, the required computational cycles or their reuse of previously learned skills.

\section{Why is gradual good?}
\label{sec:gradual}

Given a difficult task that needs to be solved, a good strategy to find a solution is to break it down into smaller problems which are easier to deal with. The same is true for learning\cite{Bengio2009-fd,Salakhutdinov2013-on}. It is much faster to learn things gradually than to try to learn a complex skill from scratch \citep{Alexander2015-mx,Amodei2016-qf,Zhang2016-ug,Zaremba2015-ia,Gulcehre2016-mb,Oquab2014-ql,Sharif_Razavian2014-ca,Azizpour2015-fd}. One example of this is the hierarchical decomposition of a task into subtasks and the \textbf{gradual learning} of skills necessary for solving each of them \citep{Krueger2009-vv}, progressing from the bottom of the hierarchy to the top\cite{Mhaskar2016-vq,Mhaskar2016-du,Poggio2015-hg} \citep{Polya2014-vb}.

\newthought{For instance}, consider a newborn child which is given the task of learning to get to the airport. The chance that the child will learn to do this is very small, as the space of possible states and actions is simply too large to explore in a reasonable amount of time. But if she is taught gradually via simpler tasks, for instance learning how to crawl first and then walk, one increases the chances of success, as the child can then exploit these previously learned skills to try to get to the airport.
\newline 

Therefore, it is beneficial to build systems which learn gradually, and to be in control of the learning process by guiding their learning in specific ways. \textbf{Guided learning} therefore means showing the system which things make sense to learn at this moment and in which order \citep{Gulcehre2016-ht,Bengio2009-fd,Vapnik2015a}. This reduces the necessity for exploration even further.

\newthought{Another benefit} of gradual learning is its generality. We do not have to specify a single global objective function (the main goal of AI) at the beginning, because we are teaching general skills, which can be used later for solving new tasks.
\newline

In the case of the child, she is initially taught to walk and open doors, even if its teachers (parents) do not yet know that she will need to get to the airport on her own, become a dentist, etc.

\newthought{If general skills} are taught gradually, the teacher might have better control over the knowledge that is learned by the system. Later, if a user specifies a goal for the system, it is more likely that in order to fulfill it, the system will try to use the already learned skills rather than invent new behavior from scratch. In this way, the chances that the system would discover an unwanted or harmful strategy could be reduced, compared to standard single-objective learning\cite{Taylor2016-bq}.
\newline 

This is similar to teaching the child how to walk and open doors, and then to go to the airport. It is more likely that the child will try to solve the task by walking and opening doors, rather than trying to learn a completely new skill (like flying) from scratch, because that would be significantly more difficult to discover or potentially even impossible.

\newthought{Gradual learning} has a number of other benefits \citep{Gulcehre2016-mb,Gulcehre2016-ht,Bengio2009-fd}\cite{Pan2010-en} over standard ``static''  learning that we believe outweigh its disadvantages (outlined in the critique of our approach in Section 9):

\begin{itemize}
\item Optimizing a model that has few parameters and gradually building up to a model with many parameters is more efficient than starting with a model that has many parameters from the beginning\cite[-1cm]{Stanley2002-nm}. At each step, you only need to optimize/learn a smaller number of new parameters\cite[-0.5cm]{Chen2015-py}.
\item There is no need for a priori knowledge of the system's architecture\cite{Zhou2012-ve,Rusu2016-pp,Ganegedara2016-cv}. Architecture size can correspond to the complexity of given problems (the architecture can start small and grow as needed\cite{Fahlman1990-sp}, in contrast to an architecture of a pre-defined final size)
\item Reuse of existing skills is feasible and even encouraged\cite{Andreas2015-av,Andreas2016-at}
\item Once a skill is acquired, it is no longer relevant how long the skill took to discover. The cost of using an existing skill is notably smaller than searching for a skill from scratch.
\end{itemize}

\chapter{How to build and educate general AI?}
\label{ch:educate}

A system capable of general AI will eventually exhibit a very large repertoire of ideally general skills. Designing such a system from scratch and learning all skills at once is infeasible. The effort is more attainable if the problem of learning and designing is deconstructed into several, less complicated ``sub-problems'' which we know how to tackle. For example, it is clear that we want the AI to understand and remember images, so it needs the ability to analyze them and a memory to store data. It is beneficial if the system is able to communicate with humans\cite{Mikolov2015}, so it will need to read, write and understand language. It will also need to learn and adapt to new things, and much more. 

\section{Skills as building blocks towards general AI}
\label{sec:skills}

Currently, our definition of skills is very broad and can encompass many concepts. Rather than individual skills, it is the gradual acquisition of skills, their interplay and the continuum of their functionality that is important.

\newthought{Skills} can range from simple general or concrete (like the ability to recognize faces, add numbers, open doors, etc.) to more complex abstract as well as specialised ones (like the ability to build a model of the world, to compress spatial and temporal data, to receive an error signal and adapt accordingly, to acquire new knowledge without forgetting older knowledge, etc.). Due to our graduality requirement, the necessity of skills to only represent general abilities is sufficient, but not necessary. Skills might also provide a way to measure how well parts of the system work, as it is clear how to measure which system is better in understanding speech, classification, and game playing \citep{Hernandez-Orallo2016}. However, it is still unclear how to evaluate general AI as a whole\cite[-6.5cm]{HernandezOrallo2017,Shieber2006-tv,Savova2007-me,Thorisson2016}.

\section{Intrinsic vs. Learned Skills}
\label{sec:intrinsic}

Some of the skills that a general AI system will possess might be \textbf{intrinsic}, i.e. hard-coded by programmers, but most will be \textbf{learned}. Take, for example, the evolution of humans\cite{Darwin1859-du}. Evolution provided us with certain very general intrinsic skills or predispositions, but most of what we know, we need to learn during our lifetimes - from our parents, the environment, or society. Those skills cannot be hard-coded because humans, just like an AI, need to be able to adapt to unknown future situations. Sometimes it is also easier to teach the desired skill through an appropriate curriculum, than to add it as a part of the design. On the other hand, letting the AI discover all skills by itself, especially very general ones, would be slow and inefficient\cite[-2cm]{Thomaz2006-ss,Brys2015-yg,Peng2016-it}. This means that our job is to identify essential skills and find efficient ways to transfer them to a general AI system - by hard-coding them or by teaching them. It is not necessary to find the best skills. Any skills which have the desired properties and which enable the AI to further learn and improve itself can move us closer to general AI.

\section{How to optimize the process?}
\label{sec:optimize}

Just like an AI has to use efficient and effective methods when searching for problem solutions, AI researchers must also look for shortcuts to narrow the search for a general AI architecture and an optimal learning curriculum, as we cannot effectively explore the entire space of all potential solutions. We can, for instance, draw inspiration from evolution \citep{Bengio2014-tz,Darwin1859-du}\cite{Kandel00}, animal brains\cite{Oh2016-ni,Toda2015-ke}, or other systems, designs and processes\cite{Nise2007-pj}. Part of the problem is also which general AI architecture and skill set is easier for us to attain now, with our current knowledge and resources.

\begin{figure*}[h]
\forcerectofloat
  \includegraphics[width=\linewidth]{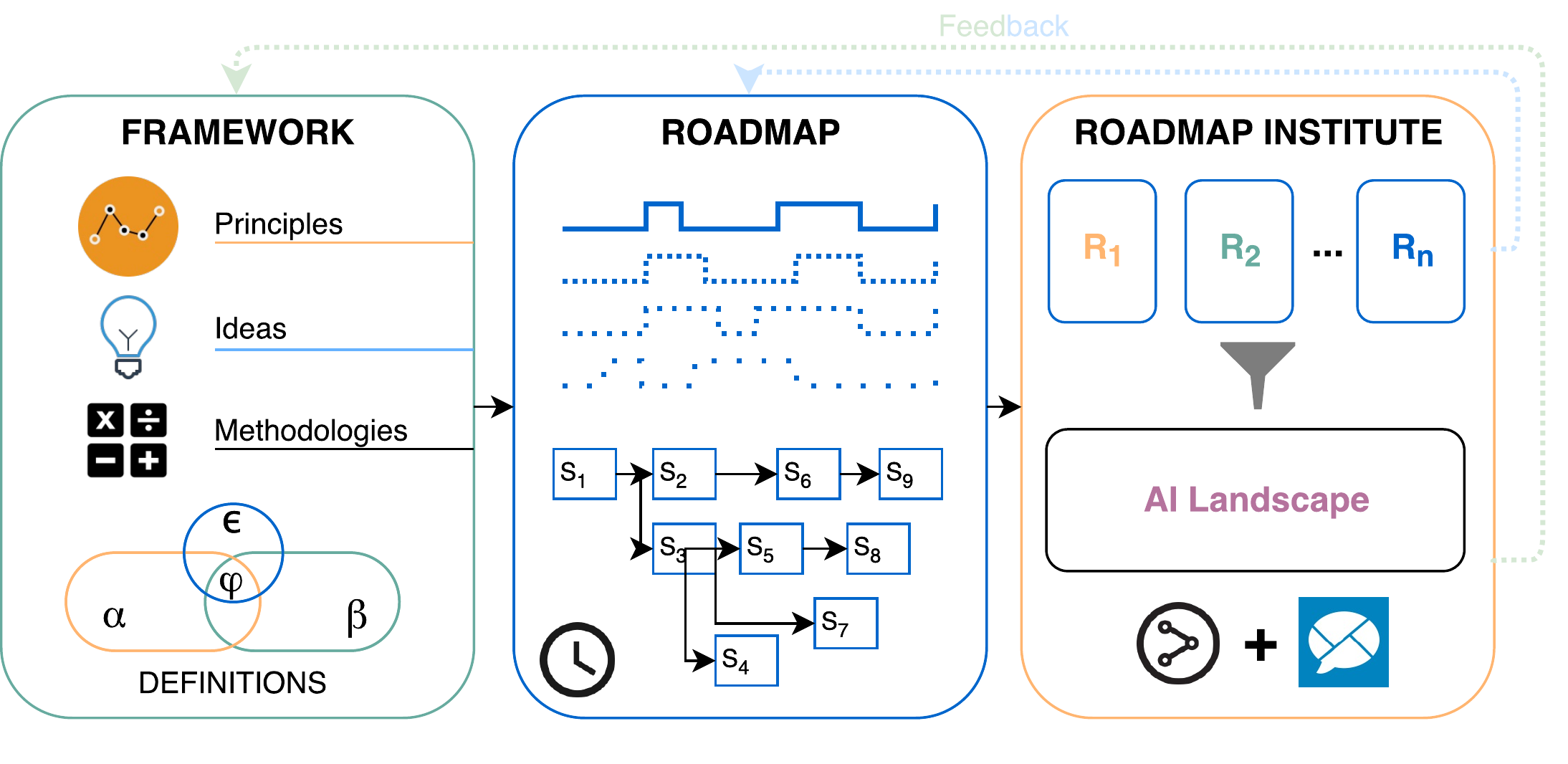}%
  \caption{Conceptual overview of our proposed approach to optimizing the process of building general AI. A \textbf{framework} defines the principles, ideas, methodologies and definitions that underlie our search for general AI. \textbf{Roadmap} defines a partially ordered set of skills ($S_i$) which are either to be learned gradually in a guided way, or already be present in the system at its inception. The \textbf{AI Roadmap Institute} analyses different roadmaps ($R_j$) and frameworks to develop an understanding of the entire field and provide continual feedback and discussion about both, frameworks and roadmaps.}%
  \label{fig:optimize}%
\end{figure*}

\newthought{This framework}, roadmap and the institute (see following sections, respectively and Fig. \ref{fig:optimize}) are all part of our approach  for optimizing the process of searching for general AI. In other words, narrowing the search for the architecture and the learning curriculum.

We can start asking questions like,``What is the minimal skill set that is sufficient for human-level general AI?'' If we can optimize the process by cutting out all unnecessary skills\cite{Blumer1990-xi}, we can get to our goal faster. On the other hand, a learning algorithm alone wouldn't be sufficient; we also need the thousands or millions of learned skills for a particular domain. Without them, the AI would not be able to start solving the problems we give it. For example, baking a cake is not likely to be a crucial skill for an AI system attempting to solve difficult medical problems, spending most of its time studying medical journals. But a skill such as the ability to generalize to similar, but previously unseen situations is universal, and falls into the category of necessary skills for every general AI. Through continual feedback during architecture creation, learning of a curriculum, analyses and comparison by the institute and collaboration with the wider community, answers to many other questions, such as ``How is our approach better compared to others?'', ``Is the notion of skill sufficiently well defined?'', ``Is the ordering structure of our curricula sufficiently rich?'' and many others can be continually provided and refined.

\chapter{The importance of a roadmap to \\ general artificial intelligence}
\label{ch:roadmap}

Our roadmap to general AI is a collection of research milestones (cf. section "\textit{Milestones}") that we deem essential for progress towards creating human-level intelligent machines. It is a separate, additional document and an essential entity\cite{Rosa2016a} to this framework. It can be seen in Figure \ref{fig:ADRoadmap} in the Appendix. Currently, we partition our roadmap into two primary areas, containing three complementary parts:

\begin{enumerate}
\item Research and Development - \textit{how to get to general AI?}

\begin{itemize}
\item \textbf{Architecture Roadmap} - \textit{intrinsic skills and architecture design}
\item \textbf{Curriculum Roadmap} - \textit{learned skills and gradual knowledge acquisition}
\end{itemize}

\item Future and Safety - \textit{what to expect next?}

\begin{itemize}
\item \textbf{Safety/Futuristic Roadmap} - \textit{how to keep humanity safe and the years leading up to and after general AI is reached}
\end{itemize}

\end{enumerate}

\begin{figure*}[h]
  \forceversofloat 
  \includegraphics[width=\linewidth]{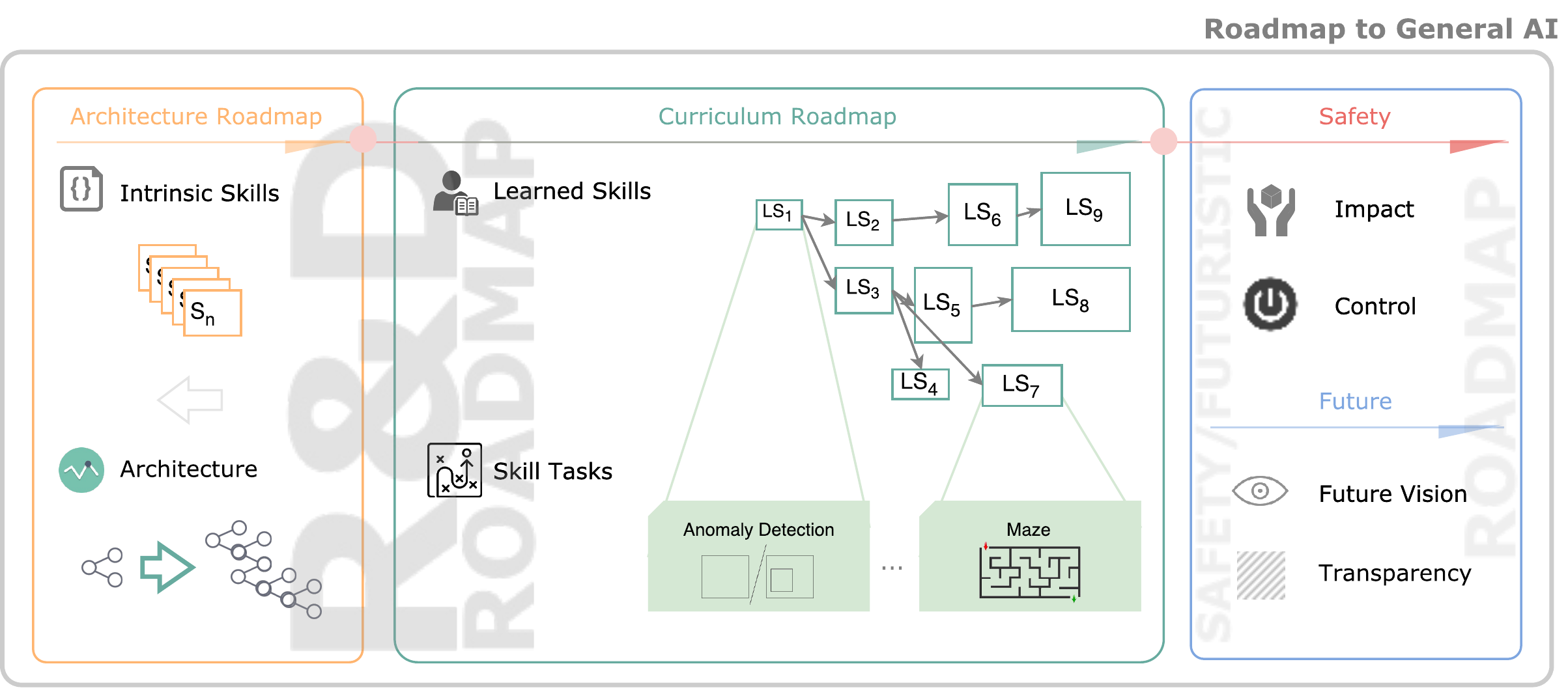}%
  \caption{An overview of our \textbf{roadmap to general AI}. The roadmap is a collection of research milestones, divided into four complementary roadmaps, each dealing with different aspects of the process of building a general AI system. The Architecture Roadmap guides the development of intrinsic skills and architectural design, the Curriculum Roadmap defines the gradual acquisition of skills, necessary to reach human-level performance, and the Safety and Futuristic Roadmap provides insights into what to expect next and how to be prepared.
}%
  \label{fig:roadmap}%
\end{figure*}

Fig. \ref{fig:roadmap} puts these into context. It provides an overview of the entire roadmap to general AI. 

\newthought{The research and development roadmap} (Architecture + Curriculum Roadmaps) are partially ordered lists of skills which our AI will need to be able to exhibit in order to achieve human-level intelligence. Ordering of skills is currently inspired by Piaget\cite{Flavell2002-jz}, but other hierarchies might also offer some additional insights\cite{Keith2010}. Each skill or a set of skills represents a solution to a possibly open research problem. These problems can be distributed among different research groups, either internally at GoodAI, or amongst external researchers and collaborators.

\newthought{To learn general skills}, we can define what tasks the agent should be able to solve. Based on tasks that have not been solved yet, we can derive a \textbf{research problem} (what researchers should figure out). Solving this research problem results in at least one of the following:

\begin{itemize}
\item a modified architecture which can solve the tasks (exhibits new intrinsic skill(s))
\item a modified architecture which can learn how to solve the tasks (exhibits new intrinsic and/or learned skill(s))
\item a modified curriculum, in which the system can learn to solve tasks (acquired new learned skill(s))
\end{itemize}

\newthought{New skills} very often depend and build upon previously acquired skills, so the research problems exhibit some intrinsic dependencies. We cannot simply skip to a skill in the middle of the roadmap and start acquiring it because this would go against the principle of reuse of previously learned skills. Such discontinuity might break the inherent dependency on previously learned skills. Instead, each skill is like a stepping stone to a subsequent skill. It is very important that an architecture that has the ability to solve problems (tasks in the roadmap) does not approach each problem in isolation \citep{Pan2010-en}. On the contrary, the solution of a problem could ideally be directly based on the solutions of previous, simpler problems \citep{Pan2010-en,Rusu2016-pp}, i.e. on previously learned skills Under such a graduality requirement, some problems that are "solved" in the traditional sense of the word (like chess or checkers) still remain open.

\newthought{Our roadmap} is a living document which will be updated as we work towards a set of milestones and evaluate them within this framework. The current version of this document is early-stage and a work in progress. We anticipate that more milestones and research directions will be entered into the roadmaps as our understanding matures, during agent learning, and ideally as a result of analyses produced by the proposed AI Roadmap Institute 
\newline 

There are still many parts missing, but we feel that it is better to engage with the community sooner rather than later [4]. The first version of the roadmap can be seen in the Appendix.

\section{Milestones}
\label{sec:milestones}

Our research roadmap describes the stages of general AI development. In summary, there are five main stages of development that we will outline here for completeness. In each of the stages (shown in Table 1) we have an environment, a teacher, and an AI system. The environment and teacher work in tandem to teach the AI a set of useful, ideally general skills. In Table 1, each stage defines the end state of a milestone.

\bigskip
\begin{table}[p!]
  \footnotesize%
  \begin{center}
    \begin{tabular}{r p{15cm}}

Table 1				& \textbf{Milestones}\\ \midrule
Stage \textbf{0} 	& \textit{Before any code is written}\\  \cmidrule(r){2-2}
					& This is a stage before AI gets a chance to start learning.
                    \newline \\
Stage \textbf{1}	& \textit{Nothing learned but some intrinsic skills present }\\ \cmidrule(r){2-2}
					& In the first stage, the AI starts with zero learned skills. It already has some intrinsic (hard-coded) skills that are very general, for example, the capacity to acquire skills in a gradual manner, the reuse of skills, a rudimentary form of recursive self-improvement, the capacity to learn through gradual and guided learning, etc. 
                    \newline
                    
                    In other words, it has the potential to learn new skills and use these new skills to improve its learning capabilities.
                    \newline\\
                    
Stage \textbf{2}	& \textit{Simple general skills learned}\\ \cmidrule(r){2-2}
					& The AI learns simple general skills through gradual and guided learning, which can be seen as, for example, a mix of puppet, supervised, apprenticeship learning or multi-objective reinforcement learning. The AI uses an error signal (feedback) from a teacher to change its behavior towards desired outcomes.
\newline

The goal is to teach the AI a set of general skills that will be useful in follow-up learning. The AI needs to learn to emulate these skills and behaviors.
\newline

Special attention is given to teaching how to communicate via a simple language, how the world works, etc., so that all skills the AI may learn in the future can already build on top of these skills, making all follow-up learning more efficient.
\newline

At this point, the AI does not need to learn any very specific knowledge (e.g. the capital of the Czech Republic, the name of the president, etc). It learns only general skills.
\newline

The AI does only a little self-exploration at this stage. Given its repertoire of skills, self-exploration would be less effective than during later stages. Our requirement for the system is to simply emulate the skill-set provided by the teacher.
\newline\\

Stage \textbf{3}		& \textit{Complex and specialized skills, language and exploration learned through indirect feedback} \\ \cmidrule(r){2-2}
						& At this stage, the AI is fully capable of communicating with the teacher and a hardwired error feedback signal will not often be needed \citep{Singh2004,Mohamed2015-zw,Machado2016-ft}, due to the possibility of providing instructions and feedback via language.
\newline

The AI has associated positive and negative feedback with messages received through language \citep{Mikolov2015}.
\newline

It keeps learning additional complex and specialised skills - reasoning, communication, etc., as well as additional useful knowledge.
\newline

The AI also learns how to efficiently explore the world on its own. Rather than being hard-coded only, exploration using a large repertoire of already learned skills should result in more meaningful, effective and efficient exploratory behavior compared to a naive policy. The more skills the system learns, the better further exploration becomes.
\newline

It also continues in self-exploration \citep{Machado2016-ft}, principally guided/biased by the skills/behaviors acquired in previous stages. 
\newline\\

Stage \textbf{4}		& \textit{Human-level general AI}\\ \cmidrule(r){2-2}
						& We have a fully developed general purpose AI that has a large repertoire of skills which are needed and can be directed toward any goal, solving as many tasks to at least the same level of accuracy as most humans \citep{Rajani2011}.
\newline

In its free time (when the AI system is not working on any particular goal from humans), it continues self-learning and knowledge consolidation in preparation for anticipated future goals from humans.
\newline

The AI continues to recursively self-improve, eventually resulting in general AI transcending human-level abilities.\\ 

\end{tabular}
\end{center}
\end{table}

\bigskip

\section{Guidelines for working with the roadmap}
\label{sec:guidelines}

In order to help guide the development of roadmaps and to provide concrete workflows for how we approached the creation of our roadmap, we will publish a detailed description of this process\cite{Andersson2016}. This "workflow" document outlines a guideline for building curricula (sequences of problems) for the education of artificial agents. It also presents guidelines for building a system that learns to solve the problems in the curricula.

\newthought{Our motivating assumptions} are that \textbf{well-designed curricula can accelerate learning} in agents as they do in humans. Furthermore, curriculum development, like software development, can benefit from being guided by a defined process. We expect that several guiding principles exist which are shared by all good implementations of incremental learning systems.

We believe that the following key questions are essential for building successful curricula:
\begin{itemize}
\item How to come up with problems that will appear in the curriculum?
\item How to determine the right time to present a problem in the curriculum?
\item How to decide that a problem is useful / not useful or is missing?
\item How to decide that two problems are similar / different?
\item How to correctly evaluate agent's solution to a problem?
\end{itemize}

The ``workflow'' document contains four major sections focusing on answering the above questions and more.

\begin{enumerate}
\item Creating problems/skills 
\item Creating curricula and using a curriculum roadmap
\item Evaluating curricula
\item Solving problems contained in the curricula
\end{enumerate}

\begin{figure}[h]
  \includegraphics[width=\linewidth]{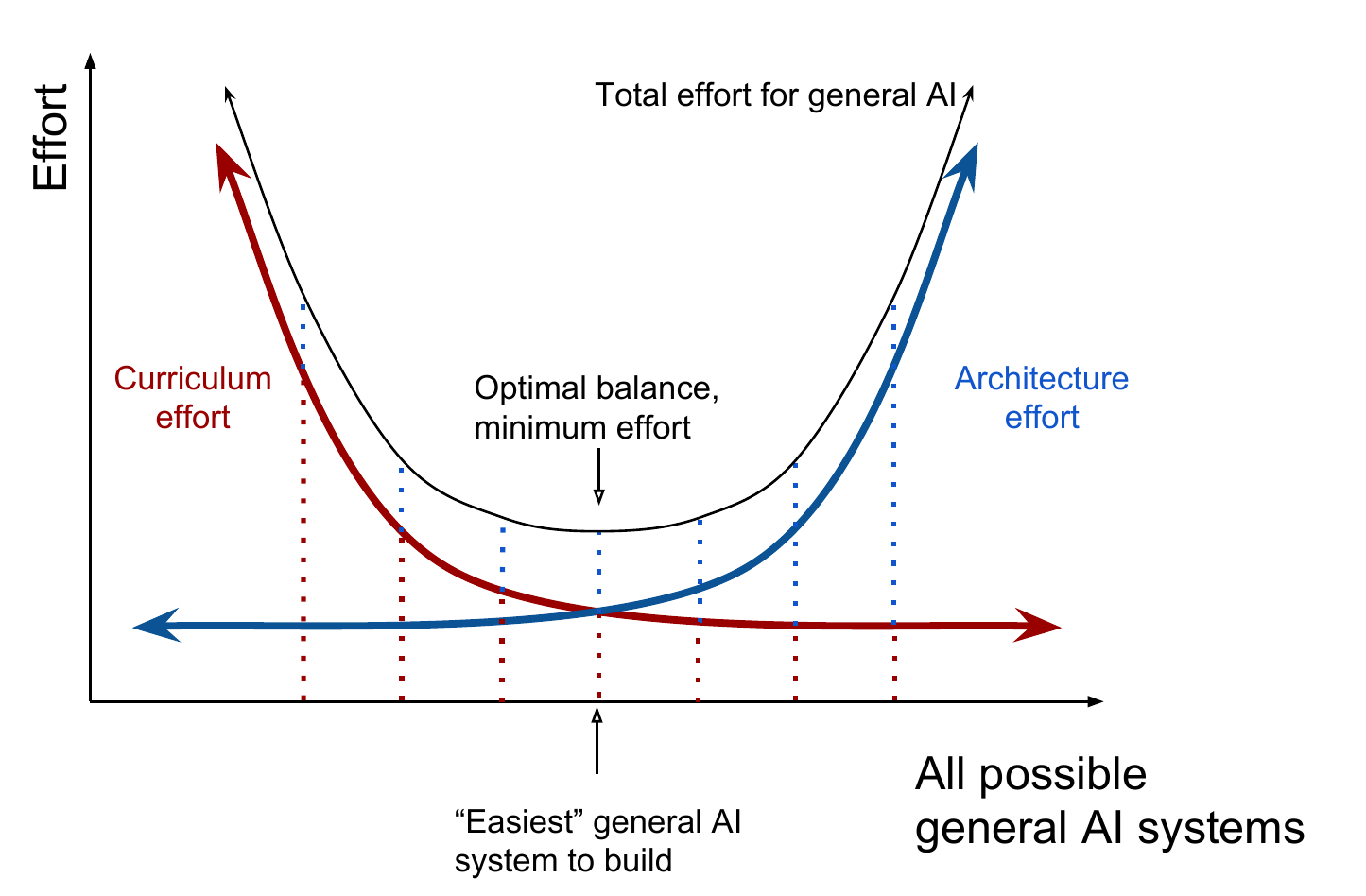}%
  \caption{Visual depiction of the trade-off between effort exerted for building architectures for a general AI vs. effort spent on developing curricula in our School for AI. Among all possible general AI systems and curricula, there exists an optimal architecture-curriculum pair that minimizes the total effort, while achieving general AI.}%
  \label{fig:balance}%
\end{figure}

Figure \ref{fig:balance} highlights the importance of the creation of good curricula matched with the right architecture. We believe that efforts spent on both of these problems are equally important and go hand in hand. The workflow document provides some insight into this process in order to minimize the time and effort spent on converging to a human-level general AI system.

\section{Futuristic / safety roadmap and framework}
\label{sec:futuristic}

In addition to our Roadmap to general AI, our Futuristic Roadmap is our vision for the future and the specific step-by-step plan we will take to get there. This roadmap outlines challenges we expect to come across in our development and efforts to keep general AI safe, and how we will mitigate risks and difficulties we will face along the way. 

\newthought{Our futuristic roadmap} is a statement of openness and transparency from GoodAI, and aims to increase cooperation and build trust within the AI community by inspiring conversation and critical thought about human-level AI technology and the future of humankind. While our R\&D roadmap is focused on the technical side of general AI development, this futuristic roadmap is focused on safety, society, the economy, freedom, the universe, ethics, people, and more.

\chapter{AI Roadmap Institute}
\label{ch:institute}

In an attempt to provide a platform for better collaboration and understanding between AI researchers, we propose the creation of an independent AI Roadmap Institute. We are founding and starting this new initiative\cite{TheGoodAICollective2016} to collate and study various AI and general AI roadmaps proposed by those working in the field, map them into a common representation and therefore enable their comparison. The institute will use architecture-agnostic common terminology provided by this framework to compare the roadmaps, allowing research groups with different internal terminologies to communicate effectively.

\newthought{The amount of research} into AI has exploded over the last few years, with many new papers appearing daily. The Institute's major output will be consolidating this research into a comprehensible visual summary which outlines the similarities and differences among roadmaps. It will identify where roadmaps branch and converge, show stages of roadmaps which need to be addressed by new research, and highlight examples of skills and testable milestones. This summary will be presented in a clear and comprehensible way to maximize its impact on as wide an audience as possible, minimizing the need for significant technical expertise, at least at its 'big picture' level.  It will be constantly updated and available for all who are interested. An overview of the Roadmap Institute, for illustrative purposes displayed using an example common representation based on our framework, can be seen in Figure \ref{fig:institute} below.

\begin{figure}[h]
  \forceversofloat 
  \includegraphics[width=\linewidth]{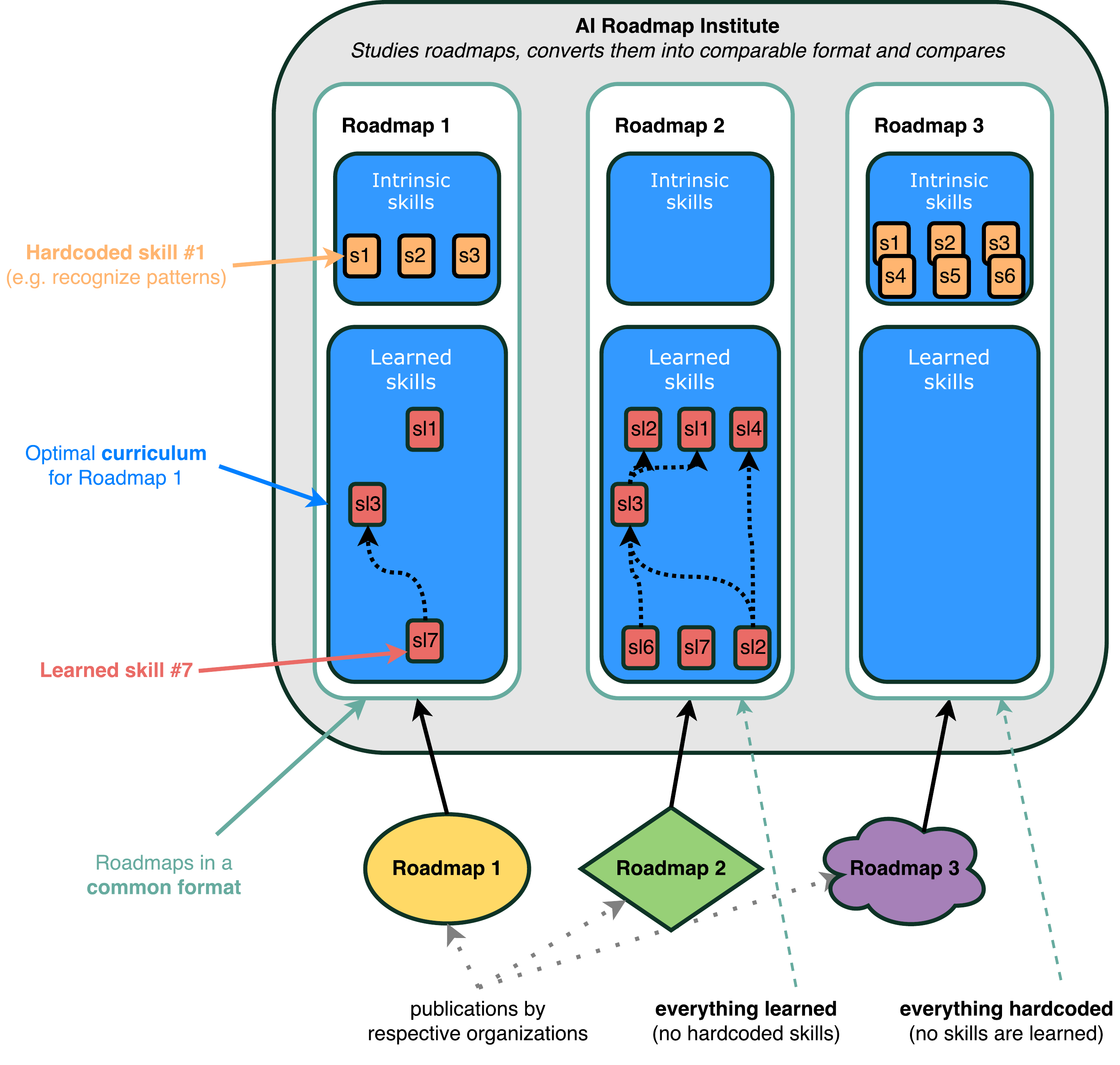}%
  \caption{Overview of the \textbf{Roadmap Institute}, showing its primary purpose of studying and understanding published roadmaps, converting them into a comparable representation and allowing for meaningful comparison and tracking of progress of the AI field as a whole. The shown representation is based on this framework. This is for illustrative purposes and one of many possibilities of how to compare roadmaps. Ultimately, it is the Institute's mission to find the best such common representation.}%
  \label{fig:institute}%
\end{figure}

These roadmaps will be described by the institute in an implementation agnostic manner. The roadmaps will show problems and any proposed solutions, and the implementations of others will be mapped out in a similar manner.

\newthought{The institute} is concerned with 'big picture' thinking, without focusing on many of the local problems in the search for general AI. With a point of comparison among different roadmaps and with links to relevant research, the institute can highlight aspects of AI development where solutions exist or are needed. This means that other research groups can take inspiration from or suggest new milestones for the roadmaps.

\newthought{Finally}, the institute is for the scientific community and everyone will be invited to contribute. It will phrase higher level concepts in an accessible and architecture-agnostic language, with more technical expressions made available to those who are interested.

\chapter{School for AI / Curriculum learning}
\label{ch:school}

Our Roadmap to general AI is the basis for curricula that define the way in which our general AI system is to learn. The realization of such curricula is what our School for AI is for. Besides having hard-coded general intrinsic skills, we naturally expect the system to be able to learn. We will teach the AI new skills in a gradual and guided way using this School for AI which we are now developing. 

\newthought{In the School for AI}, we first design an optimized set of learning tasks, or a "curriculum". The aim of the curriculum is to teach the AI system useful skills and abilities, so it does not have to discover them on its own. When the curriculum is ready, we subject the system to training. We  evaluate its performance on learning tasks of the curriculum and use it to improve the curriculum itself as well as the set of learned and intrinsic skills.

\section{Main principles}
\label{sec:schoolprinciples}

Gradual learning means learning skills incrementally, where more complex skills are based on previously learned, possibly simpler, skills. 

\newthought{Guided learning} means that there is someone (a mentor or society) who has already discovered many skills for us, and we can learn these skills from them. Guided learning is extremely important, because without it, the AI would waste time exploring areas that evolution and society have already explored, or that we know are not useful or perhaps even dangerous.

\section{Curriculum requirements}
\label{sec:curriculumrequirements}

A good curriculum: 
\begin{enumerate}
\item Minimizes the time needed for getting the general AI into a desired target state. When the AI is in a target state, it can learn and evolve on its own;
\item Is efficient - i.e. not more complex than necessary;
\item Minimizes the number of skills that need to be hard-coded into the general AI.
\end{enumerate}

\newthought{Finding the optimal curriculum} for the AI is a multi-objective optimization problem. The better the curriculum, the faster the learning. However, it might not be possible to design a universally optimal curriculum (cf. the "no free lunch theorem"\cite{Schaffer1994-gv}), yet reaching for human-level general AI might allow for avoiding this theoretical argument \citep{Hernandez-Orallo2016}. We are limited by the level of our current knowledge that we can transfer and by the eventual architecture of the general AI. 
\newline 

However, we believe that a high-quality curriculum can optimize the learning process and allow for faster breakthroughs in AI over purely algorithmic advances. 

\section{Artificial learning environment}
\label{sec:learningenvironment}

For teaching the AI, we have created a simulated visual "toy world" with simplified physical laws that is grounded in simple language. We are designing our curriculum to teach the AI from the most basic rules of the world to the most complex ones, up to the point where it can start learning on its own. 

\newthought{The goal} is not to teach the AI any arbitrary and specific facts about the world. On the contrary, it is to teach the AI useful and general skills for a more efficient understanding and exploration of the world, and for better and more general problem solving.

\newthought{During development} of the School for AI, we encountered an interesting problem - how should we specify the tasks for the AI? When there little or no common language, it is very challenging and time-consuming to explain what the AI needs to do. For this reason, we are also focusing on early language acquisition. To cut down on AI development time, we want to be able to efficiently communicate with it as soon as possible.

\chapter{Gradual learning competition}
\label{ch:competition}

As mentioned in Section \ref{ch:introduction}, we believe that one of the most fundamental challenges in developing human-level intelligent machines is the creation of agents that have the ability to acquire and reuse skills and knowledge in a gradual manner. Unlike in an unconstrained setting, this problem continues to pose serious challenges under bounded resources\cite{Gershman2015}. To truly and quickly progress in this area, we propose the injection of a monetary stimulus to the AI community in the form of a competition. We suggest launching the competition in two stages: 

\begin{enumerate}
\item \textbf{Stage 1} - Identification of requirements, specifications and a set of evaluation tasks for gradual learning
\item \textbf{Stage 2} - Development and implementation of an agent that gradually learns and passes requirements defined in stage 1
\end{enumerate}

Upon completion, the winner will receive a significant monetary prize (millions of \$), provided by us and potentially by other investors.

\chapter{Development of theoretical foundations}
\label{ch:theory}

There is nothing so practical as a good theory\footnote{Kurt Lewin}. Having described our understanding of intelligence, explained the importance of acquiring skills, and in particular learning in a gradual and guided manner, what are the theoretical foundations that underlie our ideas and proposed solutions?

\newthought{Our framework} outlined in this document is in its first "analytic" stage according to Russell's philosophical method \citep{Klement2013}. It is very general and it neither adheres to one particular theory, nor subsumes it. This allows us to investigate and formalize our thoughts through the perspective of various differing theoretical approaches in its second "constructive" phase \citep{Klement2013}. These include the following approaches which we are actively investigating:

\begin{itemize}

\item \textbf{Information Theory}
\begin{itemize}
\item Algorithmic Information Theory\cite[-6cm]{Li2013-xn,Solomonoff1995-ep,Hernandez-Orallo2015}
\item Kolmogorov Complexity \citep{Li2013-xn}
\item Minimum Description/Message Length\cite[-2cm]{Rissanen1978-hi,Wallace1968-xh}
\end{itemize}

\item \textbf{Learning Theory}
\begin{itemize}
\item Vapnik-Chervonenkis Theory\cite[-1.5cm]{Vapnik98}
\item Rademacher Complexity\cite[-1cm]{Bartlett2001-me}
\item Robust Generalization\cite{Cummings2016}
\end{itemize}

\item \textbf{Computational Mechanics} and \textbf{Statistical Physics}
\begin{itemize}
\item Structural complexity\cite{Crutchfield2011-aq}
\item $\epsilon$-machines and transducers\cite{Barnett2015-yh}
\item Integrated Information\cite{Tononi2015-yx}
\end{itemize}

\end{itemize}

\newthought{Despite} fundamental links among many of the above approaches, formalizing our concepts, ideas and the framework as a whole through a variety of theories allows us to examine and scrutinize our core concepts from different viewpoints. In this way, we hope to have a clearer picture of all the possibilities, as well as, limitations of our approaches. 

Examples of areas where we are beginning to observe benefits of such theoretical analyses:
\begin{itemize}
\item Measures of gradual accumulation of skills
\item Task and curriculum complexity measures
\item Evaluating adaptive/growing architectures
\end{itemize}

This work is ongoing and will be continually added to this framework to provide solid theoretical foundations to much of the work that we undertake.

\chapter{Related work}

Although there is a lack of unified perspectives on the building of general artificial intelligent machines, a number of works are important to mention here. Mikolov et al. proposed one of the more complete frameworks for building intelligent machines \citep{Mikolov2015}. Their approach focuses on communication and learning in a simplified 'toy' environment, similar to our School for AI, yet significantly more limited. \citet{Lake2016} on the other hand propose a more philosophical discussion of the limitations of current approaches. They argue that solving pattern recognition is not sufficient and that causal discovery is essential. Likewise, grounding models in the physical laws of the world and the exploitation of compositionality and learning-to-learn approaches are vital for rapid progress towards intelligent machines.  

\newthought{Ideas about} \textbf{learning environments} and \textbf{curricula} have been developed in a number of works. \citet{Bengio2009-fd} initiated the BabyAI project and introduced curriculum learning: learning accelerated by presenting easy examples first and progressively increasing the difficulty. In the framework of \citet{Mikolov2015}, learning in a simplified environment is also presented. In their work, however, the environment is defined by language only and no visual input is provided to the AI. An interesting discussion on the shortcomings of present AI systems and efforts at Facebook to build systems that learn for general AI is found in the work of Bordes et al.\cite{Bordes_A_Weston_J_Chopra_S_Mikolov_T_Joulin_A_Rush_S2015-qv}. Project Malmo, an interactive 3D toy environment based on the game Minecraft is discussed in by \citet{Johnson2016}\cite{Johnson2016}. \citet{Thorisson2016} argue that a theory of AI tasks can give us more rigorous ways of comparing and evaluating intelligent behavior. Stages of human cognitive development are described, for example, in \citet{Flavell2002-jz}\cite{Flavell2002-jz}. A thorough overview of evaluation environments, measures and challenges in AI is presented by \citet{Hernandez-Orallo2016}.

\newthought{A number of developments} in the field of AI and machine learning are of interest to our work. Due to the vast amount of exciting research conducted in this field in the last few years, only works that are particularly relevant are discussed. 
\newline 

For a brief \textbf{overview of  deep learning} architectures, \citet{LeCun2015-ls}\cite{LeCun2015-ls} provide a concise and high-level overview. For an in-depth analysis and reference of the field, the work of \citet{Schmidhuber2015-pn}\cite{Schmidhuber2015-pn} is recommended. 
\newline 

Of particular interest are \textbf{growing architectures} that learn new things while retaining existing knowledge. This area of research is closely related to our interest in gradual learning. Besides references provided throughout this framework, here we list additional related works. A systematic overview of a related field of 'transfer learning' is provided by \citet{Pan2010-en}. \citet{Schmidhuber2013-hr} presents a framework for automatically discovering problems inspired by playful behavior in animals and humans. \citet{Rusu2016-pp}\cite{Rusu2016-pp} introduce progressive neural networks that adapt to new tasks by growing new columns. In earlier work, \citet{Fahlman1990-sp} demonstrated accelerated learning by adding one hidden neuron at the time, keeping the preceding hidden weights frozen. Similar additive learning capabilities are demonstrated for convolutional neural networks by \citet{Li2016}\cite{Li2016}. \citet{Nivel2013} prototype "a machine that becomes increasingly better at behaving in under specified circumstances, in a goal-directed way, on the job, by modeling itself and its environment as experience accumulates". \citet{Steunebrink2016-dd} introduce Experience-based AI, a class of systems capable of continuous self-improvement. Architectures that can learn as much as possible of their structure from training data are also relevant for creating progressively growing agents. \citet{Zhou2012-ve} introduce an algorithm for learning features incrementally and determining architecture complexity from data. In contrast, rather than using a simple heuristic, \citet{Cortes2016-cy}\cite{Cortes2016-cy} propose a growing neural network architecture algorithm exploiting theoretical bounds derived through a statistical learning complexity measure. \citet{Saxena2016-ob}\cite{Saxena2016-ob} propose a convolutional neural fabric that learns the structure of convolutional networks. Gradual learning in a toy environment and exploiting reinforcement learning can be found in the work of \citet{Oh2016-ni}.

\newthought{Another important area} is \textbf{compositional learning}, i.e. the ability to form knowledge about a particular subject by unifying knowledge about multiple other subjects that are already understood. \citet{Vincent2008-pg}\cite[-2cm]{Vincent2008-pg} describe the use of denoising autoencoders to improve the representational power of deep networks by composition. \citet{Andreas2016-at} compose neural networks from component networks for a visual question answering task. 
\newline 

\textbf{Learning programs} or algorithms are another form of compositional learning. Such methods allow an agent to represent and reuse procedural knowledge. \citet{Reed2016} introduce neural programmer-interpreters able to compose hard-coded instructions into programs\cite{Reed2016}. Similarly, \citet{Neelakantan2015-bl} train an agent to perform table lookups on data using a number of intrinsic operators\cite{Neelakantan2015-bl}. \citet{Kaiser2015-wy}'s neural GPU learns algorithms in a network that is wide rather than deep\cite{Kaiser2015-wy}; the parallelism makes for easier training and more efficient execution. \citet{Riedel2016-rj} incorporates prior procedural knowledge as Forth program sketches with slots that can be filled with learned behavior\cite{Riedel2016-rj}.

\newthought{Other works}, not immediately relevant to the above topics but worth mentioning, are works that we believe are currently relevant to our progress towards general AI. Many works inspired by biology, namely neuroscience\cite{Kumaran2016-qb}, are of great interest. For example, spike-timing dependent plasticity (STDP) is a biologically inspired approach with the potential to improve unsupervised learning\cite{Bengio2016-tq,Scellier2016-ae,Bengio2014-tz}. \citet{Osogami2015-tl} use it to learn temporal patterns with Boltzmann machines\cite{Osogami2015-tl}. \citet{Izhikevich2007-om} addresses the problem of delayed reward in biological and artificial neural networks. \citet{George2008-gc} proposes a model of learning and recognition where temporal patterns are central. \citet{Wang2003-dc} provides an overview of approaches to the representation and processing of temporal patterns.

\chapter{Critical review}

This section is an overview of some of the most relevant critiques of our framework, roadmap, ideas and approaches. This is an ongoing list, providing an active list of research problems that must be answered in order to support or refute our ideas and approach presented in this framework.

\section{Our own assessment of risks and disadvantages of our framework}

\begin{itemize}
\item \textbf{Development Complexity} - As highlighted in Figure \ref{fig:balance}, there is a trade-off between the complexity of an architecture and the amount of effort necessary for developing a successful curriculum. If we cannot find the optimal balance within this curriculum-architecture continuum, the approach could be exceedingly demanding on our side, for programmers, researchers as well as for the development of School for AI. Heavy supervision, non-sparse feedback or sub-optimal biasing of the AI system, could result in decreasing efficiency of learning due to the infeasible demand on manpower required, for example, for supervision.

\item \textbf{Development Bias} - Gradual learning and objective-less search might result in biasing the architecture by what we teach it and in which order. If not incorporated sensibly, this could lead to sub-optimal solutions in cases when we could have learned a better skill directly. In our gradual learning example, we teach the child how to walk because we don't know how to fly. But this does not mean that flying is impossible. On the contrary, flying could be a more efficient skill for the child, yet it might not discover such skill because it will simply walk every time.
\end{itemize}

\section{Assumptions}

\begin{itemize}
\item Skills in our roadmap, that remain "open problems", may be more difficult to distribute to the scientific community than we think. Traditionally, an open problem is defined as-is, standalone, without dependencies. In our case, however, we define as an open problem issues that may have been solved already, but not in a gradual and guided way. To find a solution (for an arbitrary problem in the roadmap) that works in a gradual and guided way, we might need to give to an external researcher our current AI implementation, so that improvements could be made in order to solve the open problem.
\item Similarly, solutions that will come from our research problems may not be easy to integrate by others. Unless a common representation for solutions exists, arbitrary solutions might be impossible to compare and combine.
\item The assumption of eventual removal of reward/feedback at later stages of the development of the AI system is currently lacking any form of guarantees that will ensure that the AI will continue its expected operation and not descend into chaos, or simply into a failure-mode, unable to solve any meaningful tasks. Intrinsically motivated methods for reinforcement learning investigate this problem to some degree\cite{Singh2004}, but more insight is needed. 
\end{itemize}

\section{Problems}
\begin{itemize}
\item Gradual learning is a very difficult problem that shares many of the same issues that exist in transfer, lifelong and incremental learning\cite{Pan2010-en,Oh2016-ni,Rusu2016-pp}. Among others, these include problems related to the transfer of existing knowledge within models as well as in between parts, namely, what, when and how to transfer both learned and potential knowledge in a practically feasible manner.
\end{itemize}

\section{Limitations}
\begin{itemize}
\item Without working theoretical formalizations of our framework, it might be difficult to obtain various necessary quantitative measures, such as task complexity measures, determination of optimal gradual skill acquisition, evaluation criteria and others.
\item Interplay between agent and curriculum development - there are many questions and issues that should be addressed:
\begin{itemize}
\item Autonomous exploration - What if differently structured agents discover features of the world / skills in different order?
\item Impact of modifying a curriculum as we learn more about how the agent learns
\item Avoiding the "incestuous circle" if we modify the curriculum. Will multiple agent development streams help here?
\item Strive for a single optimal curricula, or multiple satisfactory ones?
\end{itemize}
\end{itemize}

\section{Performance}

\begin{itemize}
\item There are some potential drawbacks of gradual learning from the computational efficiency point of view. For example, the fact that we do not know the size of an architecture a-priori limits us in the ability to efficiently optimize and exploit such a predefined space. This could also result in inefficient "growing" of an architecture that is unable to reuse parts of already learned knowledge that is not rooted in a suitable part of the solution space.
\end{itemize}

\section{Generality}

\begin{itemize}
\item By imposing the necessity to acquire any skills that reduce the search space of solutions, we are inherently biasing our AI. This is beneficial when such bias has positive impact, for example in terms of data efficiency, i.e. faster learning / convergence. However, other undesirable biases might be introduced into the system. These could affect the development of our AI further down the line. This is another tradeoff that might be impossible to avoid completely, thus better control over which biases are introduced is desirable.
\item There is a tradeoff between the amount and types of intrinsic skills (and therefore speed of initial learning) and universality of the resulting architecture.
\item This framework and the associated roadmaps are our way of approaching the challenge of tackling development of general AI. Our thoughts and ideas might not be compatible enough with the ideas of others. In this case cooperation and collaboration may be limited, unless we adapt and adjust to some degree, based on feedback from the community.
\end{itemize}

\chapter{Next steps}

In this section we suggest a number of near-term plans for general AI development that we deem important and will undertake next. These might not be the optimal set of next steps, however, they are a starting point from which we can build upon. We believe these should include:

\begin{enumerate}
\item \textbf{Framework and R\&D Roadmap}
  \begin{enumerate}[label=(\alph*)]
  \item More research groups, institutes and companies should publish their own frameworks and R\&D roadmaps, in the spirit of what we propose in our framework, to encourage innovation, openness and progress
  \item \textbf{Framework}:
    \begin{enumerate}[label=(\roman*)]
    \item Should be continually updated and refined 
    \item Act as an internal as well as external research trace
    \item Increasingly provide multiple theoretical viewpoints on underlying ideas
    \end{enumerate}
  \item \textbf{Roadmap}:
    \begin{enumerate}[label=(\roman*)]
    \item Additional skills should be continually added to cover yet unmapped areas
	\item Task theory should be developed to provide foundations for curricula, including evaluation
    \end{enumerate}
  \end{enumerate}

\item \textbf{Architecture Groups }
  \begin{enumerate}[label=(\alph*)]
  \item Collaborate with researchers in Curriculum Groups
  \item Successfully implement architectures that support "gradual accumulation of skills"
  \item Test promising architectures on a subset of learning tasks specified in R\&D roadmap
  \end{enumerate}

\item \textbf{Curriculum Groups}
  \begin{enumerate}[label=(\alph*)]
  \item More learning tasks should be continually added (training and testing)
  \end{enumerate}

\item \textbf{AI Roadmap Institute} (Q1 2017)
  \begin{enumerate}[label=(\alph*)]
  \item Publish first version of comparison of roadmaps
  \item Start collaborative process of adding more roadmaps to the comparison
  \item Raise awareness about AI roadmapping and 'big picture' thinking in AI\cite{Rosa2016b}
  \end{enumerate}

\item \textbf{Open problems} specified in our (or AI Roadmap Institute) roadmap
  \begin{enumerate}[label=(\alph*)]
  \item Outsource implementation / solutions
  \end{enumerate}

\item \textbf{Gradual learning competition} (Launching Q1-Q2 2017)
\end{enumerate}

\chapter{Contributions}

There is a significant lack of unified approaches to building general-purpose intelligent machines. Comparable to the natural sciences\cite{Ledford2015}, most researchers, universities and institutes still operate within a very narrow field of focus\cite{Porter2009-vc}, frequently without consideration for the 'big picture'. 

We believe that our approach is one possible way of stepping out of this cycle and provide a fresh, unified perspective on building machines that learn to think. We hope to achieve this in a number of ways, each of which are equally relevant and essential for tackling different aspects of the building process:

\begin{itemize}
\item Our framework provides a unified collection of principles, ideas, definitions and formalizations of our thoughts on the process of developing general AI. This allows us to consolidate all that we believe is important to define as a basis on which we and possibly others can build. This is an ongoing and open process and feedback from the community will be invaluable for further refinement and standardization. Ultimately, it could act as a common language that everyone can understand, and provide a starting point for a platform for further discussion and evolution of ideas relevant for building general AI.
\item Our roadmap is a principled approach to clearly outlining and defining a step-by-step guide for obtaining all skills that a human level intelligent machine will eventually need to possess. This includes their definitions, as well as the gradual order and way in which to achieve them through curricula of our 'School for AI'. 
\item Our School for AI provides learning curricula - a principled, gradual and guided way of teaching a machine. This approach differs significantly from current approaches of narrowly focused, fixed datasets. We believe that gradual and guided learning are essential parts of data-efficient learning that are paramount to quick convergence towards a level of intelligence that is above current standards.
\item To compare and contrast existing approaches and roadmaps and foster more effective distillation of knowledge about the process of building intelligent machines, our AI Roadmap Institute is a step towards an impartial research organization advancing the search for an optimal protocol for achieving general artificial intelligence.
\item To tackle one of the most fundamental challenges in developing human-level intelligent machines, we propose the creation of a gradual learning competition. We believe, this monetary-driven stimulus could provide a further boost for the community to push the limits of our understanding of this challenging topic.
\end{itemize}

Using a language that is consonant with our principles, the above are simply a set of skills for steering our search for general AI that we believe are important and will help us achieve significantly faster convergence towards developing truly intelligent machines.

\chapter{APPENDIX}

\section{Proposed team structure}
Our ideal team is organized into five smaller working groups: Research Group, School Group, Software Engineers, Roadmapping Group and AI Safety team.

\begin{itemize}
\item The research group's focus is on the implementation of solutions to research topics, mostly focusing on growing topologies, modular networks, and the reuse of skills. Intrinsic skills are developed and implemented as part of this group. 
\item The school group studies the skills that an AI needs to learn, and designs learning tasks for efficient education. The group also works on the R\&D roadmap, mapping various curricula. This team will teach the AI learned skills. 
\item The software engineers are responsible for the entire software infrastructure for both research and development. They handle developing novel frameworks and libraries as necessary and as required by other teams, supporting significant novel forms of computation and problem handling. 
\item The internal roadmapping group studies various roadmaps developed internally as well as by the wider community. It has a general oversight of the landscape of the entire field of developing general AI and of the overall internal research process. It investigates and maps methods for combining and comparing new progress in the field.
\item The AI Safety team studies the safe path forward with our technology, and the mitigation of threats to our team and humankind as a whole. This team is creating an alliance of AI researchers committed to the safe development of AI and general AI, our futuristic roadmap, and more. 
\end{itemize}

Despite the need for close-knit collaboration among these groups, some research problems and developmental stages can be outsourced or distributed among a number of collaborators. Both the School and Safety groups are perfect examples of groups that can be extended to included external collaborators, research groups and institutions, and from which collaboration could foster stronger results.

\begin{fullwidth}
\section{Informal Definitions}

\bigskip
\begin{table}[h]
\small
  \begin{center}
    \begin{tabular}{r p{14cm}}
\toprule
Table A1			& \textbf{Definition}\\ \midrule
\textbf{Framework} 	& \textit{A \textbf{unified collection} of principles, ideas, definitions and formalizations that are believed essential for developing human-level general Artificial Intelligence}\\  \cmidrule(r){2-2}
					& This document is an example of a framework. It is our attempt at unifying all the concepts, definitions and processes that we believe are necessary for the successful development of an intelligent machine. 
\newline \\
\textbf{Roadmap}	& \textit{A \textbf{principled approach} for defining and outlining a \textbf{step-by-step guide} for obtaining all \textbf{skills} that a human-level general AI will need to possess}\\ \cmidrule(r){2-2}
					& A roadmap defines a partially ordered list of tasks that allow for an agent to learn or acquire, in a gradual and guided way, skills that are related and of increasingly higher complexity. Some are already solved problems in the AI community, while others are open problems. One example is our R\&D roadmap [62]. 
\newline \\
\textbf{AI Roadmap Institute}	& \textit{A platform for mutual \textbf{understanding} and \textbf{collaboration} within the AI landscape}\\ \cmidrule(r){2-2}
					& An initiative to collate and study the field of AI from a holistic perspective, mapping progress into common representations and allowing for easier overview (ideally visual), comparison and improved efficiency and progress in development of AI, fostering collaboration throughout.
\newline \\
\textbf{School for AI}		& \textit{Our realization and a \textbf{grounding of curricula} defined in a roadmap} \\ \cmidrule(r){2-2}
						& An optimized set of learning tasks, termed "curriculum" according to which an agent can learn useful skills in a gradual and guided way. Feedback from agent allows continual refinement of curriculum and allows for the search for an optimal trade-off between architectural and curriculum complexity.
\newline \\
\bottomrule
\end{tabular}
\end{center}
\end{table}

\bigskip

\bigskip
\begin{table}[p!]
\small
  \begin{center}
    \begin{tabular}{r p{14cm}}
\toprule
Table A2					& \textbf{Definition}\\ \midrule
\textbf{Intelligence}		& \textit{A problem-solving tool in complex, dynamic and uncertain environments}\\ \cmidrule(r){2-2}
						& Assuming that all problems can be posed as search and optimization problems, the goal of intelligence is to find the best available solution, given time and resource constraints. This usually requires narrowing the search space by the use of suitable skills.
\newline \\
\textbf{Artificial Intelligence} & \textit{A program that is able to learn, adapt, be creative and solve problems}\\ \cmidrule(r){2-2}
(system)				& Universal term encompassing both ``narrow'' (weak) and ``general'' (strong) AI, their difference being in the universality of the underlying technology.
\newline
\textbf{Narrow AI} - able to solve only very limited set of specific problems
\newline
\textbf{General AI} - aims for a human-level ability to solve problems
\newline \\
\textbf{Human-Level} (AI)	& \textit{System performing at least at the same level as most humans}\\ \cmidrule(r){2-2}
						& An ability of a machine to solve as many of the same tasks to at least the same level of accuracy as most humans \citep{Rajani2011}.
\newline \\
\textbf{Skill}		& \textit{A mechanism for narrowing down solution search space}\\ \cmidrule(r){2-2}
						& A skill is any mechanism that helps in the quest for solving a problem. From the point of view of mathematical optimization, it can be thought of as any assumption, approximation or a heuristic that reduces or simplifies the search space of all possible solutions to a given problem.
\newline \\
\textbf{Intrinsic Skill}		& \textit{A hard-coded ability that solves a general class of problems}\\ \cmidrule(r){2-2}
						& Unlike learned skills, intrinsic skills are explicitly coded into an AI system by its creator. This allows for a repertoire of minimal skill set that is necessary for the AI to progress and develop further through gradual learning.
\newline \\
\textbf{Learned Skill}		& \textit{An ability learned from experience (from data)}\\ \cmidrule(r){2-2}
						& Skills that are learned through the experience of solving tasks. The process of solving a task enables an AI system to learn the necessary mappings between the input (domain of the problem) and the output (solution of the task).
\newline \\
\textbf{Learning Task}		& \textit{A problem whose solution enables or helps verify the acquisition of a skill(s)}\\ \cmidrule(r){2-2}
						& In a curriculum, each skill is assumed to be attainable and testable through the successful solution of an associated learning task. e.g.
\newline
\textbf{Skill} - anomaly detection
\newline
\textbf{Task} - detection of a structure not conforming to the expected one
\newline \\
\textbf{Curriculum}		& \textit{A partially ordered list of learning milestones (skills and tasks)}\\ \cmidrule(r){2-2}
						& A learning curriculum comprises of a partially ordered list of skills that an AI needs to acquire. Measuring the quality of a solution of tasks associated with a skill in a curriculum allows for one type of evaluation of the level of intelligence an AI system has acquired.
\newline \\
\textbf{Gradual Learning} 		& \textit{Progressive learning of skills. Complex skills exploiting learned knowledge.}\\ \cmidrule(r){2-2}
(Curriculum Learning)			& Learning of skills one by one, where complex skills are based on previously learned skills. Learning curricula offer partially ordered lists of skills that are to be acquired in the predefined order. This order is due to the skill's increasing complexity and interdependence.
\newline \\
\textbf{Guided Learning}		& \textit{Directed acquisition of knowledge by an intelligent entity}\\ \cmidrule(r){2-2}
						& Guided learning, akin to shaping in childhood development, allows for direct control of the direction in which an AI system will gradually develop. Guidance allows for speeding up of the development of an AI system due to  intrinsic transfer of knowledge that is present in instructions passed from a teacher to the AI system through guidance instructions.
\newline \\
\end{tabular}
\end{center}
\end{table}
\end{fullwidth}

\begin{center}
\begin{figure*}[p!]
\centering
  \forceversofloat 
  \includegraphics[width=\linewidth]{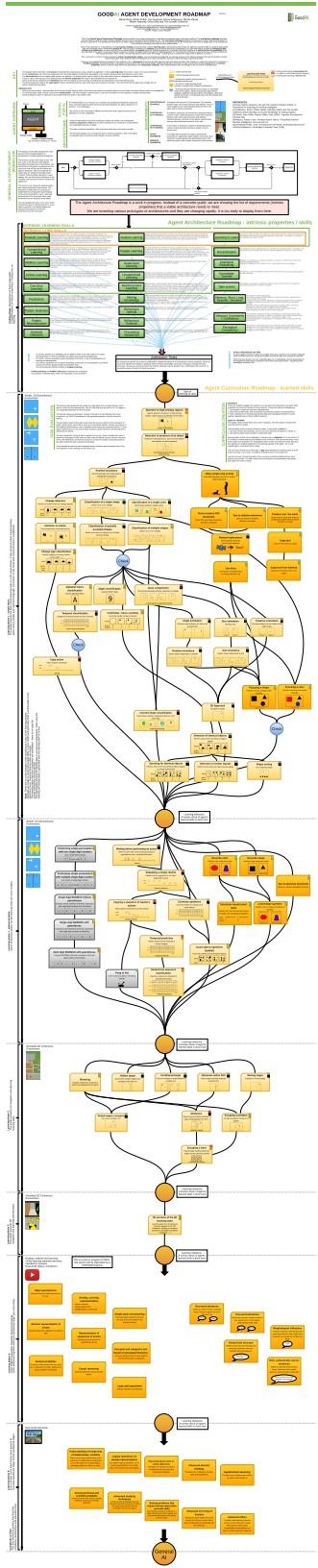}%
  \caption{\textbf{Example Roadmap}: best viewed in color, on screen and with the option to zoom in.}
  \label{fig:ADRoadmap}%
\end{figure*}
\end{center}

\backmatter

\bibliography{references}
\bibliographystyle{plainnat}


\end{document}